%% file: conference_101719.tex
\documentclass[conference]{IEEEtran}
\IEEEoverridecommandlockouts
\usepackage{cite}
\usepackage{amsmath,amssymb,amsfonts}
\usepackage{algorithmic}
\usepackage{graphicx}
\usepackage{textcomp}
\usepackage{xcolor}

\usepackage{algorithm}
\usepackage{url}
\usepackage{hyperref}
\usepackage[caption=false,font=normalsize,labelfont=sf,textfont=sf]{subfig}
\usepackage{soul}
\usepackage{multirow}
\usepackage{svg}
\usepackage{balance}

\def\BibTeX{{\rm B\kern-.05em{\sc i\kern-.025em b}\kern-.08em
    T\kern-.1667em\lower.7ex\hbox{E}\kern-.125emX}}
\begin{document}

\makeatletter
\newcommand{\linebreakand}{%
  \end{@IEEEauthorhalign}%
  \hfill\mbox{}\par%
  \mbox{}\hfill\begin{@IEEEauthorhalign}}
\makeatother

\title{Neural Edge Histogram Descriptors for Underwater Acoustic Target Recognition\\

\thanks{DISTRIBUTION STATEMENT A. Approved for public release. Distribution is unlimited. This material is based upon work supported by the Under Secretary of Defense for Research and Engineering under Air Force Contract No. FA8702-15-D-0001. Any opinions, findings, conclusions or recommendations expressed in this material are those of the author(s) and do not necessarily reflect the views of the Under Secretary of Defense for Research and Engineering. Code is publicly available at \protect\url{https://github.com/Advanced-Vision-and-Learning-Lab/NEHD_UATR}.}}

\author{
  \IEEEauthorblockN{\hspace*{-10mm} 1\textsuperscript{st} Atharva Agashe}%
  \IEEEauthorblockA{\hspace*{-10mm}
    \textit{Department of Electrical and Computer Engineering}\\
    \hspace*{-10mm}\textit{Texas A\&M University}\\
    \hspace*{-10mm}College Station, TX, USA\\
    \hspace*{-10mm}atharvagashe22@tamu.edu}%
  \and
  \IEEEauthorblockN{\hspace*{6mm}2\textsuperscript{nd} Davelle Carreiro}%
  \IEEEauthorblockA{%
  \hspace*{6mm}%
    \textit{Massachusetts Institute of Technology Lincoln Laboratory}\\
    \hspace*{6mm} Lexington, MA, USA\\
    \hspace*{6mm} davelle.carreiro@ll.mit.edu}%
  \linebreakand
  \IEEEauthorblockN{\hspace*{-15mm}3\textsuperscript{rd} Alexandra Van Dine}%
  \IEEEauthorblockA{\hspace*{-15mm}
    \textit{Massachusetts Institute of Technology Lincoln Laboratory}\\
    \hspace*{-15mm} Lexington, MA, USA\\
    \hspace*{-15mm} alexandra.vandine@ll.mit.edu}%
  \and
  \IEEEauthorblockN{\hspace*{1mm} 4\textsuperscript{th} Joshua Peeples}%
  \IEEEauthorblockA{\hspace*{2mm}
    \textit{Department of Electrical and Computer Engineering}\\
    \hspace*{2mm} \textit{Texas A\&M University}\\
    \hspace*{2mm} College Station, TX, USA\\
    \hspace*{2mm} jpeeples@tamu.edu}%
}

\maketitle
\begin{abstract}

Numerous maritime applications rely on the ability to recognize acoustic targets using passive sonar. While there is a growing reliance on pre-trained models for classification tasks, these models often require extensive computational resources and may not perform optimally when transferred to new domains due to dataset variations. To address these challenges, this work adapts the neural edge histogram descriptors (NEHD) method originally developed for image classification, to classify passive sonar signals. We conduct a comprehensive evaluation of statistical and structural texture features, demonstrating that their combination achieves competitive performance with large pre-trained models. The proposed NEHD-based approach offers a lightweight and efficient solution for underwater target recognition, significantly reducing computational costs while maintaining accuracy.
\end{abstract}

\begin{IEEEkeywords}
Texture analysis, statistical textures, structural textures, edge detection, histogram, deep learning, pre-trained models.
\end{IEEEkeywords}

\input{sections/introduction}

\input{sections/method}

\input{sections/experiments}

\input{sections/results}
 
\input{sections/conclusion}

 \section*{Acknowledgment}
Portions of this research were conducted with the advanced computing resources provided by Texas A\&M High Performance Research Computing. 

\balance
\bibliographystyle{IEEEtran}
\bibliography{references.bib}

\vfill

\end{document}

%% file: sections/introduction.tex
\section{Introduction}
\IEEEPARstart{U}{nderwater} acoustic target recognition (UATR) is crucial for applications such as environmental monitoring, exploration, and ship noise characterization, aiding in marine resource management and ocean-based technologies to enhance ocean monitoring\cite{Bjorno2017Chapter14,Maranda2008Handbook}. Passive sonar uses external acoustic signals to identify underwater objects without emitting sound \cite{Maranda2008Handbook}. Spectrograms, generated through signal processing techniques like Short-Time Fourier Transform (STFT) and Mel-frequency spectrograms, transform signals into visual representations, facilitating complex pattern extraction from acoustic data \cite{French2007Spectrograms, Ghoraani2011Time-Frequency, Yang2020Underwater}. 
\begin{figure}[!t]
\centering
\includegraphics[width=3.3in]{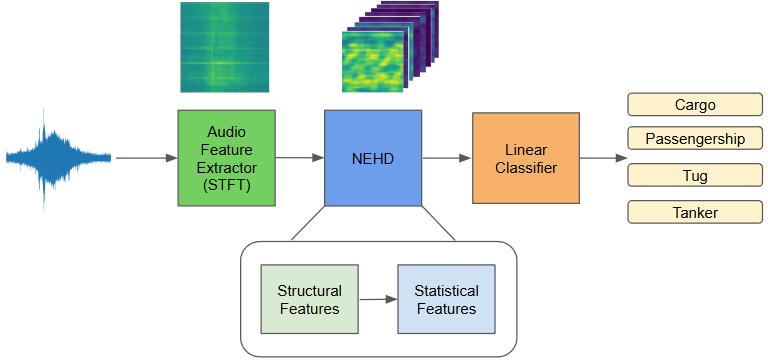}
\caption{The workflow of the proposed model for passive sonar audio classification is shown in (a). The short-time Fourier transform (STFT) is applied to the input audio signal. The neural edge histogram descriptor (NEHD) module is then used to extract structural and statistical texture features. The resulting features are then used for vessel classification.}
\label{nehd_block_diag}
\end{figure}

Classification methods for underwater acoustic signals have evolved from classic machine learning algorithms like support vector machines (SVMs) \cite{Dennis2011Spectrogram, Lu2003Content-based} to advanced deep learning techniques \cite{McLoughlin2017Continuous, Juncheng2017AComparison, Dhanalakshmi2009Classification}. While SVMs are effective, they lack feature abstraction and temporal modeling capabilities \cite{Juncheng2017AComparison}. Deep learning models, such as convolutional neural networks (CNNs), excel in feature representation and transfer learning, adapting well to underwater acoustics when pre-trained on large vision datasets \cite{Ren2019Spec-resnet, Sankupellay2018Bird, Tsalera2021Comparison}. Similarly, pre-trained audio neural networks (PANNs) \cite{Kong2020PANNs}, trained on a large audio dataset (AudioSet \cite{Gemmeke2017Audio}), have proven effective for passive sonar classification where data scarcity is a challenge \cite{Mohammadi2024Transfer}. Moreover, transformer-based models, including vision transformers (ViTs) \cite{Dosovitskiy2020AnImage} and audio spectrogram transformers (ASTs) \cite{Gong2021Ast}, have also gained prominence for their ability to model long-range dependencies, yielding state-of-the-art results in audio classification.

Texture analysis quantifies patterns and structures within images, extracting local and global features relevant to acoustic signals \cite{Armi2019Texture}. Handcrafted texture features like local binary patterns (LBP) \cite{ Ojala1994Performance} and histograms of oriented gradients (HOG) \cite{Dalal2005Histograms} are used for scene classification and sound recognition, enhancing classification performance across various acoustic tasks \cite{Ellis2011Classifying, Costa2012Music,  Abidin2018Local, Kobayashi2014Acoustic, Rakotomamonjy2015Histogram}. While pre-trained models are widely used, domain gaps limit their performance when adapted to new tasks, leading to inefficiencies and challenges \cite{Ding2023Parameter-efficient, Gong2023What}. Inspired by the efficacy of handcrafted features, this study aims to show that simpler models, which focus on extracting rich texture representations from spectrograms, can achieve performance comparable to larger models.

Neural edge histogram descriptors (NEHD) \cite{Peeples2024Histogram}, as shown in Fig. \ref{nehd_block_diag}, capture structural and statistical texture features from images through edge descriptors and a learnable histogram layer. This work applies NEHD to passive sonar data classification, demonstrating the ability to extract critical temporal and frequency features for distinguishing passive sonar signal classes. The results show that NEHD matches the performance of large pre-trained models with greater efficiency. Key contributions include: 1) evaluation of statistical and structural texture features for improved audio classification, 2) application of NEHD for passive sonar classification, achieving performance comparable to large models, and 3) use of NEHD as a feature extractor for pretrained models.

\label{sec:introduction_v2}

%% file: sections/method.tex
\section{Method}

\subsection{Overview of Neural Edge Histogram Descriptor}

NEHD is a feature extraction method designed to capture both structural and statistical texture information from an input. Following notation from \cite{Peeples2024Histogram}, in Equation \ref{deqn_nehd1}, NEHD consists of two key functions: 

\begin{equation}
\label{deqn_nehd1}
f(\mathbf{X}) = \phi \left( \sum_{\rho \in \mathcal{N}} \psi(x_\rho) \right)
\end{equation}

\noindent $\psi$ and $\phi$, that extract structural and statistical texture features respectively. $\mathbf{X}$ represents feature maps and $\mathcal{N}$ represents the local neighborhood. NEHD can be applied to input feature maps or images, such as STFT spectrograms generated from passive sonar data. The resulting texture feature maps can be used for classification.

\subsection{Structural Texture Features: Edge Descriptor Layer}

The first component of NEHD focuses on extracting structural texture features from the input spectrogram using edge descriptors. This operation, denoted as \( \psi \), applies learnable filters over a local neighborhood to capture edge information, which is necessary for highlighting the contours and boundaries present in the time-frequency representation of the spectrogram \cite{Marmanis2018Classification, Hussein2012Spectrogram}. The edge filters can be randomly initialized or set to standard edge kernels such as Sobel filters.  

The edge descriptors operation is formally described by Equation \ref{deqn_nehd2}:

\begin{equation}
\label{deqn_nehd2}
\psi_{bk}^{NEHD} = \sum_{m=1}^{M} \sum_{n=1}^{N} w_{mnk} x_{r+s+m, c+t+n, k}
\end{equation}

\noindent where \( \psi_{bk}^{NEHD} \) represents the output of the edge descriptors layer for the \( k \)-th feature channel. In this equation, the input spectrogram values \( x_{r+s, c+t, k} \) are weighted by the kernel weights \( w_{mnk} \), where \( M \times N \) defines the kernel size, and \( (r, c) \) represents the spatial dimensions of the input.

\subsection{Statistical Texture Features: Histogram Layer}

After extracting structural features through edge descriptors, the next step in NEHD involves capturing statistical texture information using a histogram layer \cite{Peeples2022Histogram}, denoted as \( \phi \). The histogram operation aggregates structural texture features from the edge descriptors layer across a sliding window and produces normalized frequency counts, capturing the distribution of texture features. The operation of the histogram layer is defined by Equation \ref{deqn_nehd3}:

\begin{equation}
\label{deqn_nehd3}
\phi_{rcbk} = \frac{1}{ST} \sum_{s=1}^{S} \sum_{t=1}^{T} \exp \left( -\gamma_{bk}^2 \left( \sum_{\rho \in \mathcal{N}} \psi_{bk}(x_\rho) - \mu_{bk} \right)^2 \right)
\end{equation}

\noindent where \( \phi_{rcbk} \) represents the output of the histogram layer for the \( k \)-th channel, and \( \psi_{bk}(x_{\rho}) \) is the structural texture response from the edge descriptors layer. The histogram operation aggregates the structural features over a window of size \( S \times T \), producing normalized frequency counts based on the exponential weighting of the deviation from the learned bin centers, $\mu_{bk}$, and bin widths, $\gamma_{bk}$.

\subsection{NEHD Model Architecture}

\begin{figure}[h!]
\centering
\includegraphics[width=3.3in]{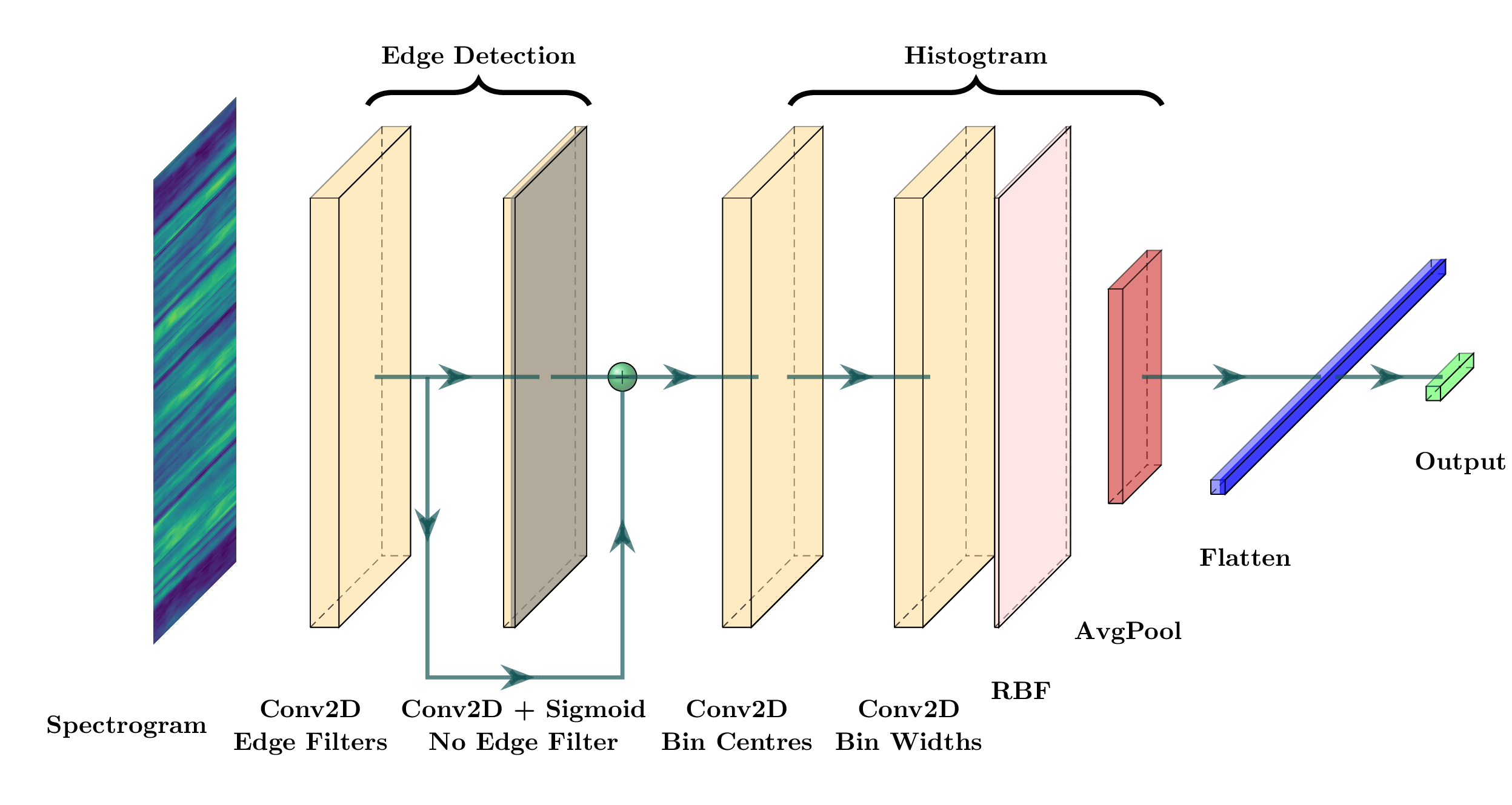}
\caption{NEHD model architecture is shown. The input spectrogram first has structural and statistical features extracted through the edge descriptors and histogram layers respectively. These features are then used for classification.}
\label{nehd_arch}
\end{figure}

\begin{figure*}[htbp]
\centering
\subfloat[]{\includegraphics[width=0.31\textwidth]{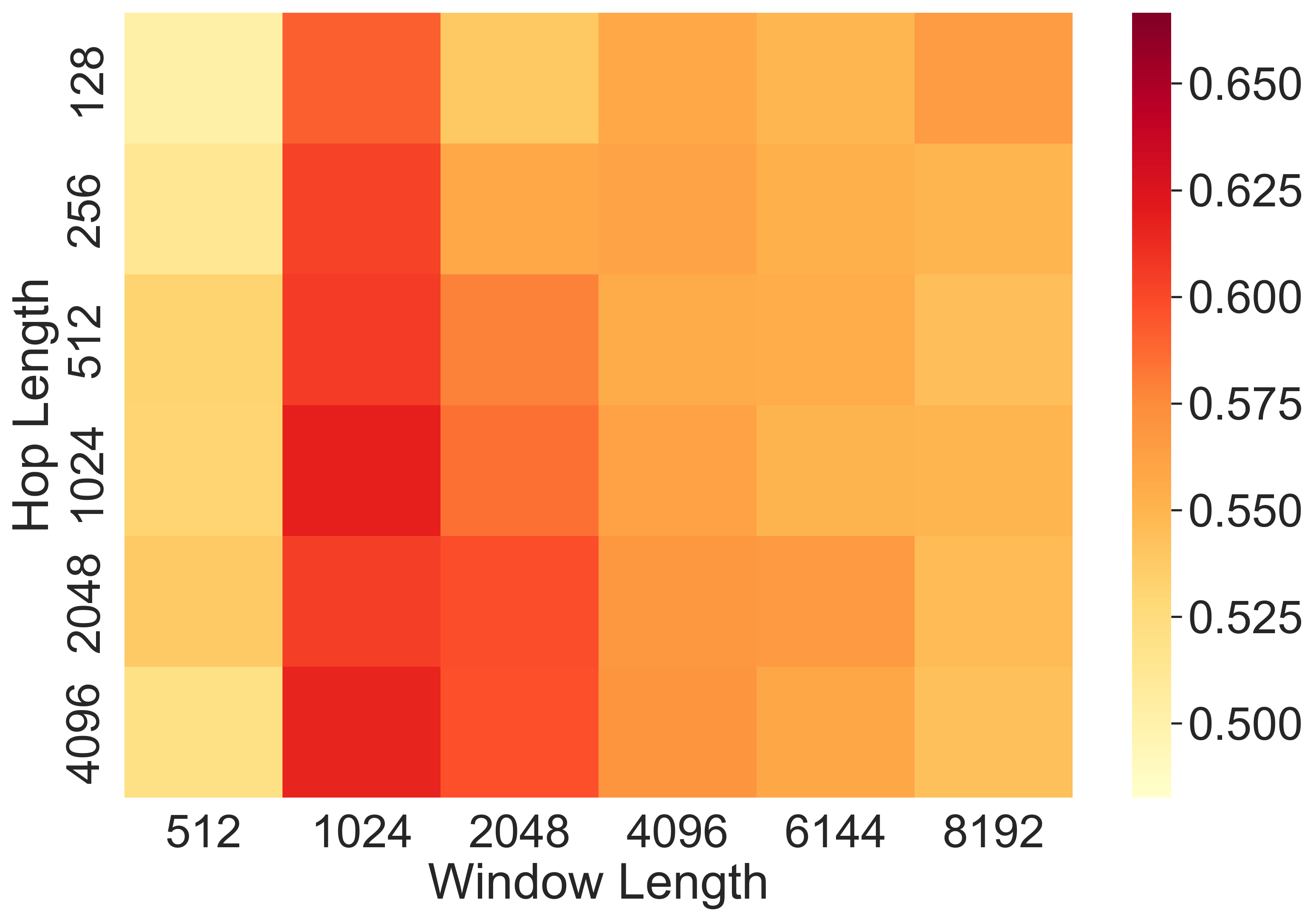}%
\label{48_cm}}
\hspace{0.02\textwidth} 
\subfloat[]{\includegraphics[width=0.31\textwidth]{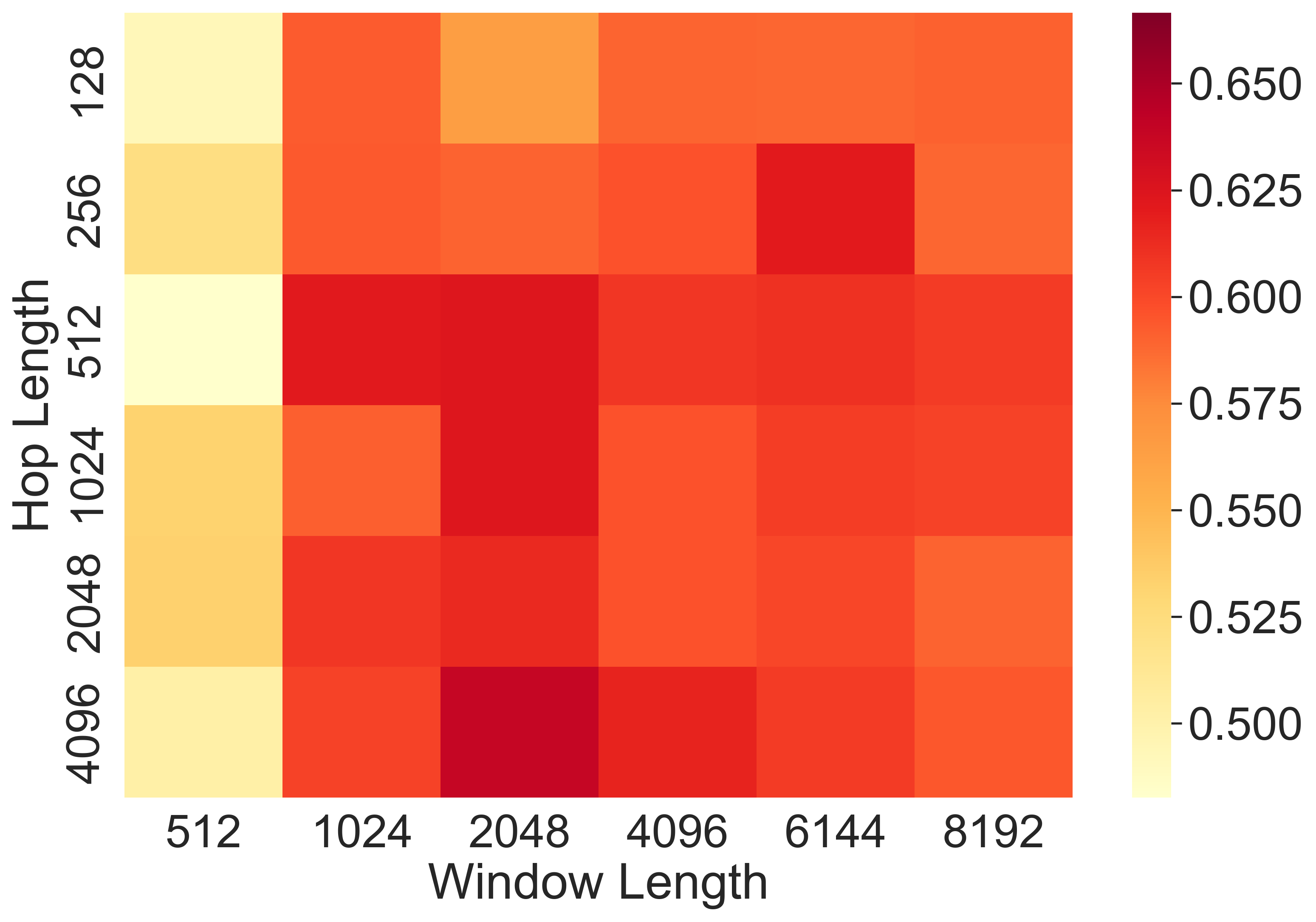}%
\label{96_cm}}
\hspace{0.02\textwidth} 
\subfloat[]{\includegraphics[width=0.31\textwidth]{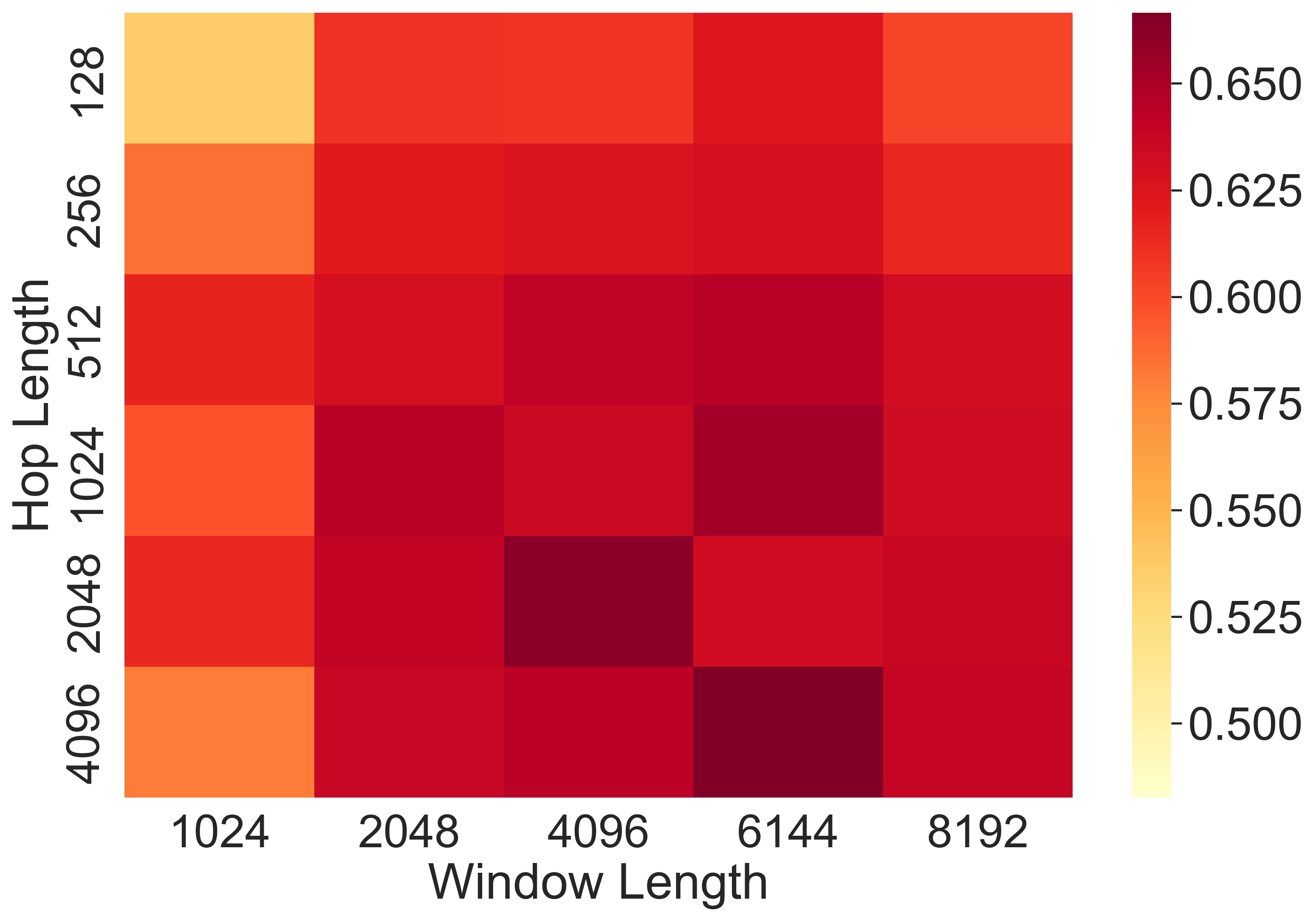}%
\label{192_cm}}
\caption{The heatmaps illustrate the exhaustive feature search process used to determine the optimal window length, hop length, and frequency bin parameters for the STFT feature. Each heatmap corresponds to a specific number of frequency bins: (a) 48, (b) 96, and (c) 192. The Y-axis represents the hop length, while the X-axis shows the window length. Darker shades indicate higher accuracy.}
\label{stft_select_cm}
\end{figure*}
The NEHD architecture as shown in Fig.\ref{nehd_arch} consists of two layers: an edge descriptors layer and a histogram layer. The edge descriptors layer comprises two blocks. The spectrogram is first processed through the convolution block (edge filters), which produces features corresponding to the number of filters. The resultant features are processed by another convolution layer consisting of a single filter and a sigmoid function, which acts as a threshold. This produces a feature map without the edges. Subsequently, the features from the edge and no-edge blocks are concatenated by a residual connection resulting in a total of edges + 1 feature maps. The resulting feature maps are subsequently processed by the histogram layer comprised of two convolution blocks for learning the bin centers and bin widths respectively followed by a radial basis function (RBF). The output of these blocks is passed through a pooling layer. This results in feature maps corresponding to the number of bins. In this case, the number of bins equals the number of edges.

\label{sec:method}

%% file: sections/experiments.tex
\section{Experiments}


\subsection{Dataset}
The DeepShip dataset \cite{Irfan2021DeepShip} used for this work is comprised of four classes: Cargo, Passengership, Tug and Tanker. The dataset was split into training, validation, and test sets using a 70-10-20 ratio, resulting in 425 signals for training, 62 signals for validation, and 122 signals for testing. Each signal was segmented into 3-second windows following previous work \cite{Irfan2021DeepShip,Ritu2023Histogram}. The ``binning" of these signals yielded 40,718 samples for training, 5,901 samples for validation, and 9,971 samples for testing.

\begin{figure}[t]
\centering
\includegraphics[width=3.3in]{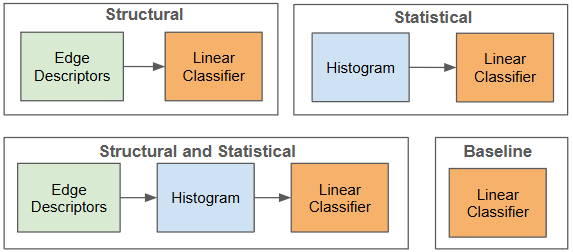}
\caption{Block diagrams illustrating individual models used to evaluate the effect of statistical vs. structural texture features.}
\label{statistical_v_structural_bd}
\end{figure}

\subsection{Experimental Procedure and Setup}
The evaluation of NEHD for passive sonar included a hyperparameter search for optimal spectrogram features, an analysis of structural vs. statistical texture impacts, a comparison with large pretrained models, and an exploration of NEHD as a complementary feature extractor with STFT for audio classification.
Two categories of models were selected for benchmarking: pre-trained ImageNet and audio models. A CNN and transformer model were selected for pre-trained model types. For the ImageNet models, ResNet-50 \cite{He2015Deep} and ViT-small \cite{Dosovitskiy2020AnImage} were used whereas for the pre-trained audio models, CNN14 \cite{Kong2020PANNs} and Audio Spectrogram Transformer (AST) \cite{Gong2021Ast} were used. 

All benchmarks were conducted over three runs with random initialization, using Adam for 50 epochs, a batch size of 128, and a patience setting of 40. The learning rate was set at 0.001 for experiments in subsections \ref{sect:NEHD_Feat_sel} and \ref{sect:NEHD_strct_stat}. In subsections \ref{sect:NEHD_benchmark} and \ref{sect:NEHD_Feat}, a learning rate of 0.001 was used for NEHD, while pretrained models were trained with a learning rate of 0.0001. In subsection \ref{sect:NEHD_Feat}, varying learning rates were applied to each respective model. The models were implemented PyTorch Lightning 2.3.0, and nnAudio 0.3.3 \cite{Cheuk2020nnAudio} was used for audio processing. The experiments were performed on an NVIDIA A100 GPU.

\label{sec:experiments}

%% file: sections/results.tex
\section{Results and Discussion}

\begin{figure*}[htb]
\centering



\subfloat[]{%
\includegraphics[width=0.24\textwidth]{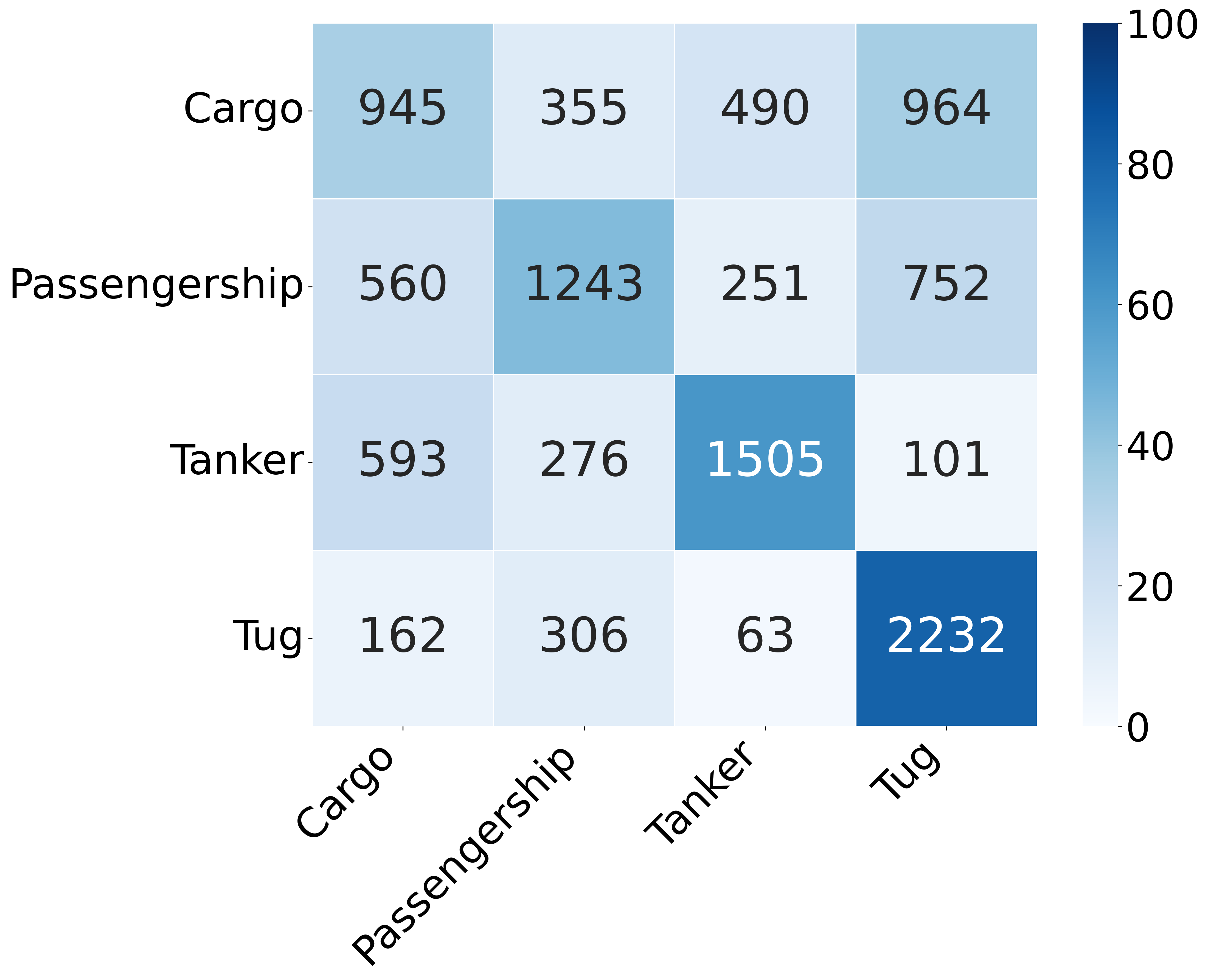}%
\label{linear_cm}}
\hfil
\subfloat[]{%
\includegraphics[width=0.24\textwidth]{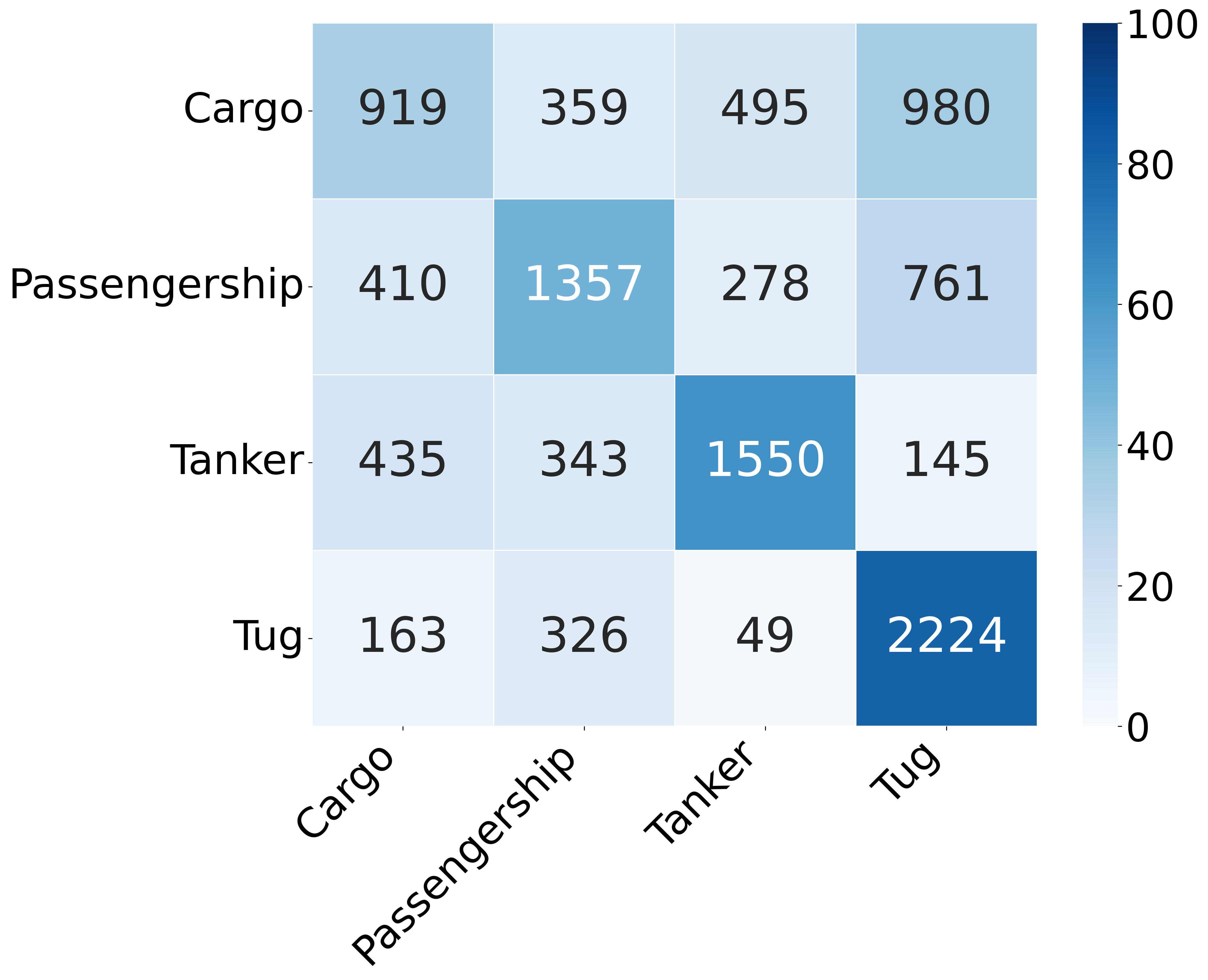}%
\label{ed_cm}}
\hfil
\subfloat[]{%
\includegraphics[width=0.24\textwidth]{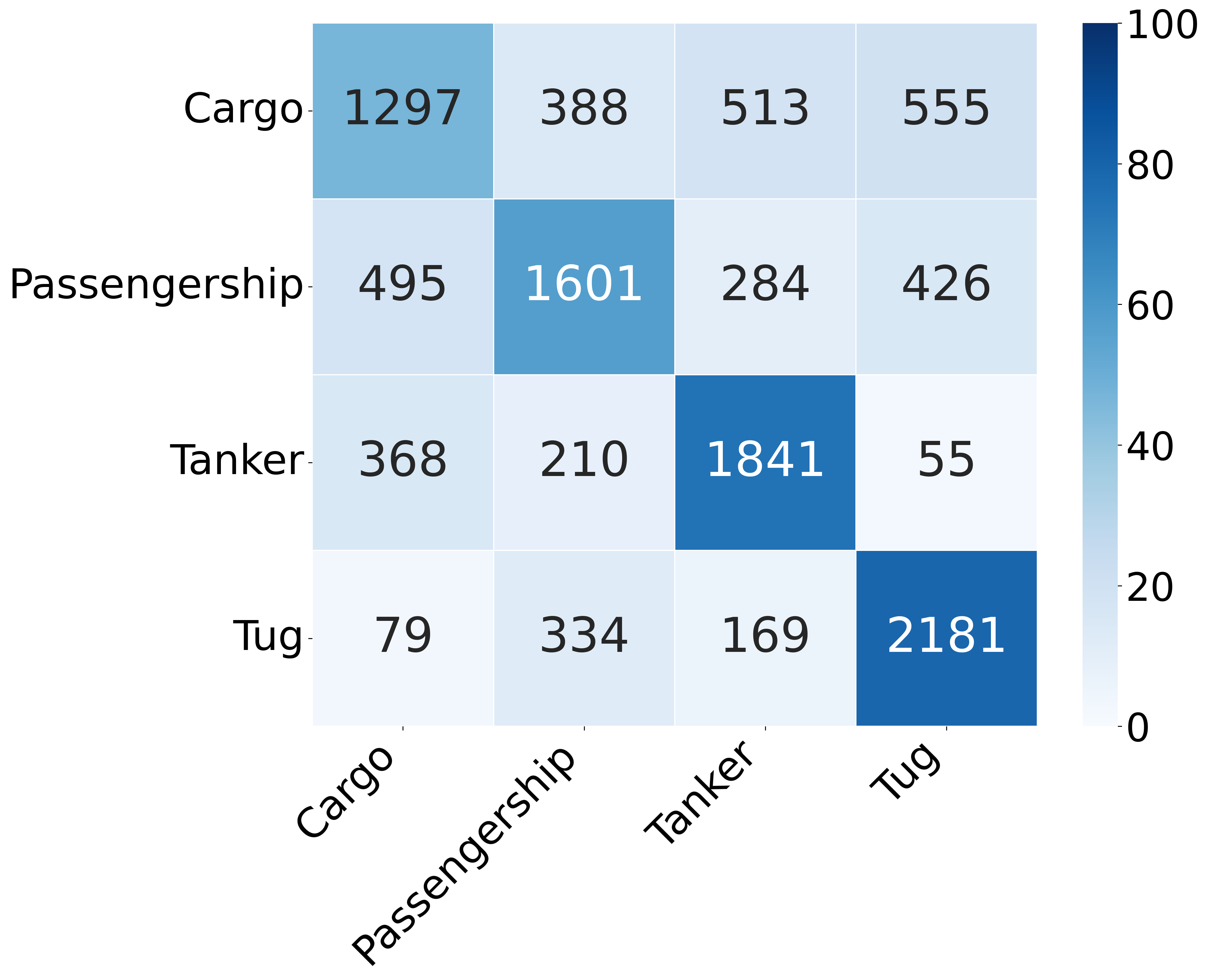}%
\label{hist_cm}}
\hfil
\subfloat[]{%
\includegraphics[width=0.24\textwidth]{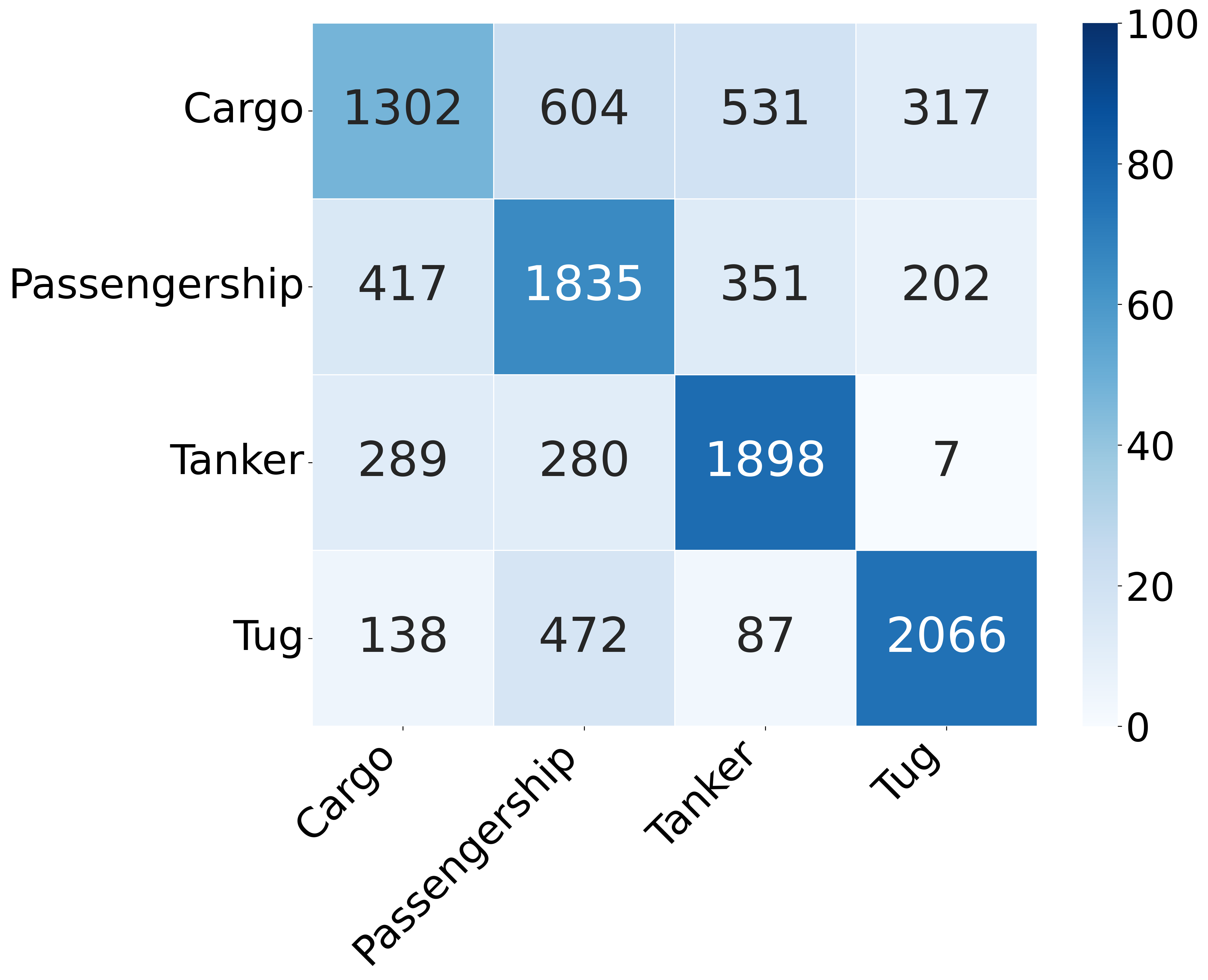}%
\label{nehd_cm}}

\caption{The average confusion matrices for (a) Linear Classifier (Baseline), (b) Edge Descriptors (Structural), (c) Histogram (Statistical) and (d) NEHD (Both Structural and Statistical) are shown for the DeepShip dataset. Darker shades indicate higher average accuracy. The predicted classes are shown along rows and the true labels are shown along the columns. NEHD improves the identification of each class in comparison to other models.}
\label{stat_struct_cm}
\end{figure*}

\subsection{Feature Selection and Extraction}
\label{sect:NEHD_Feat_sel}
 The STFT feature is used for its simplicity and effectiveness in capturing the time-frequency characteristics of acoustic signals. Although STFT captures stationary content within segments, this feature remains an efficient approach for analyzing the temporal and frequency features of acoustic signals \cite{Jahromi2019Feature}. To optimize the performance of the STFT feature, a comprehensive study was conducted involving various combinations of window lengths, hop lengths, and frequency bins with NEHD shown in Fig. \ref{stft_select_cm}. The parameter configuration that yielded the best results consisted of a window length of 6144, a hop length of 4096, and 192 frequency bins. This configuration produced features with dimensions of 192 (frequency bins) by 12 (time frames) and was used for the remainder of the experiments.

\subsection{ Effect of Structural vs. Statistical Texture Features}
\label{sect:NEHD_strct_stat}
\begin{table}[htb]
\renewcommand{\arraystretch}{1.3}    
\caption{Average test accuracy with $\pm 1$ standard deviation for 
comparison of Statistical and Structural Textures features. The best average accuracy is bolded. }
\label{tab1}
\centering
\begin{tabular}{| c | c | c | c |}
\hline
Method & Statistical & Structural & Accuracy \\ \hline
Linear Classifier &  &  & 54.89$\pm$1.15 \\ \hline
Edge Descriptors &  & \checkmark & 56.06$\pm$3.09 \\ \hline
Histogram Layer & \checkmark &  & 64.12$\pm$1.37 \\ \hline
NEHD & \checkmark & \checkmark & \textbf{65.80$\pm$0.41} \\ \hline
\end{tabular}
\end{table}

In this experiment, the performance of four models was evaluated: (a) Linear Classifier (Baseline), (b) Edge Descriptors (Structural), (c) Histogram (Statistical) and (d) NEHD (Both Structural and Statistical)
, as depicted in Fig. \ref{statistical_v_structural_bd}. The objective was to assess the effectiveness of structural and statistical features individually and their combined impact with NEHD. Table \ref{tab1} shows the evaluation metrics summary, and the confusion matrices for each model are shown in Fig. \ref{stat_struct_cm}

 The baseline linear model demonstrated a bias towards the Tug class, performing well there but poorly in other categories. The ED model outperformed the baseline by incorporating structural texture features, improving separation between Tug and Tanker classes but struggling with Cargo and Passengership classes. The histogram model enhanced accuracy for Cargo and Tanker by capturing frequency distribution patterns, yet it fell short for Passengership, suggesting the limitations of statistical features alone. The NEHD model, which integrates structural and statistical elements, achieved the highest performance by combining the strengths of ED and Histogram approaches, improving classification for all classes and demonstrating the advantage of using both texture feature types.

\subsection{NEHD vs. Pretrained Models}
\label{sect:NEHD_benchmark}
\begin{table}[htb]
\renewcommand{\arraystretch}{1.3}    
\caption{Average test accuracy with $\pm 1$ standard deviation of different pre-trained models and nehd along with the number of parameters across three experimental runs. The best average accuracy is bolded.}
\label{tab2}
\centering
\begin{tabular}{| c | c | c |}
\hline
Model & Accuracy & Parameters \\ \hline
ResNet50 & 65.63$\pm$0.46 & 23.5M \\ \hline
ViT & 64.17$\pm$0.49 & 21.5M \\ \hline
PANN & \textbf{69.92$\pm$1.00} & 79.7M \\ \hline
AST & 66.86$\pm$0.90 & 85.3M \\ \hline
NEHD & 65.80$\pm$0.41 & 13.6K \\ \hline
\end{tabular}
\end{table}

The results in Table \ref{tab2} show that the proposed NEHD model achieves an accuracy of 65.8\%, which is comparable to ResNet-50 (65.63\%) and slightly higher than ViT (64.17\%), while exhibiting lower variance ($\pm$0.41), indicating consistent performance across training runs. Although PANN achieves the highest accuracy (69.92\%), it requires a substantial parameter count of 79.7 million, which is far greater than NEHD's 13.6 thousand parameters. Similarly, AST attains an accuracy of 66.86\%, yet it also has a high parameter count of 85.3 M. The compact architecture of NEHD, with approximately only 0.016\% of the parameters of PANN and AST, showcases its efficiency and suitability for resource-constrained environments. 

\subsection{NEHD as a Feature Extractor} 
\label{sect:NEHD_Feat}

\begin{table}[htb]
\renewcommand{\arraystretch}{1.3}
\caption{Average test accuracy with $\pm 1$ standard deviation of stft versus combination of stft and nehd as a feature extractor across three experimental runs. The best average accuracy for each model is bolded.}
\label{nehdf}
\centering
\begin{tabular}{| c | c | c |}
\hline
Model & STFT & STFT+NEHD \\ \hline
ResNet50 & 65.63$\pm$0.46 & \textbf{66.03$\pm$1.64} \\ \hline
ViT & 64.17$\pm$0.49 & \textbf{65.27$\pm$0.65} \\ \hline
PANN & \textbf{69.92$\pm$1.00} & 68.77$\pm$1.15 \\ \hline
AST & 66.86$\pm$0.90 & \textbf{67.30$\pm$0.82} \\ \hline
\end{tabular}
\end{table}

The results in Table \ref{nehdf} show that incorporating NEHD as a feature extractor enhances performance while adding only 171 additional parameters. For ResNet-50, ViT, and AST, including NEHD improves accuracy, while in the case of PANN, there is a slight decrease in performance. One possible explanation is that models pretrained on ImageNet are exposed to diverse categories, including textures, objects, and patterns. ImageNet-trained CNNs, like ResNet-50, are particularly biased towards recognizing textures  \cite{Geirhos2022ImageNet-trained}, allowing them to effectively leverage NEHD's texture-based features and generalize beyond their primary domain. Additionally, transformers like ViT and AST are more robust than PANN, a CNN-based model, when integrating NEHD features, as they efficiently manage diverse tasks and minimize performance loss when handling irrelevant tasks \cite{Zhou2021ConvNets}. Overall, the results show NEHD's effectiveness as a feature extractor, enhancing model performance with minimal computational overhead.

\label{sec:results}

%% file: sections/conclusion.tex
\section{Conclusion}
These experiments underscore the importance of combining structural and statistical texture features, positioning NEHD as a bridge between traditional feature extraction and deep learning. The effectiveness of NEHD, originally designed for image data, is demonstrated for UATR, achieving classification performance comparable to deep learning models like ResNet-50, ViT, and AST while drastically reducing the parameter count. With only 13.6K parameters, NEHD is approximately 5,800 times smaller than larger models like PANN and AST, highlighting its computational efficiency and suitability for resource-constrained environments. Moreover, NEHD effectively complements STFT features and improves classification performance. This lightweight approach provides a practical solution for applications where large models are not feasible. Future work could further optimize the NEHD architecture or integrate lightweight texture extractors, such as histograms of gradients or edge filters like Prewitt, Canny, and Gabor, to enhance performance and adaptability for broader audio classification scenarios.

\label{sec:conclusion}